\documentclass[letterpaper, 10 pt, conference]{ieeeconf}  % Comment this line out if you need a4paper
\IEEEoverridecommandlockouts
\usepackage{graphicx} % Required for inserting images
\usepackage{hyperref}
\usepackage{amsmath}
\usepackage{amssymb}
\usepackage{comment}
\usepackage{stfloats} % Allow full-width figures at the bottom of a page
\usepackage{tikz}
\usetikzlibrary{arrows.meta}

\title{\LARGE \bf Co-design of trajectory and morphology for a vertical jump-climbing robot}

\author{{Christopher Y. Xu$^1$, Elliot W. Hawkes$^1$}
\thanks{$^{1}$Department of Mechanical Engineering at the University of California Santa Barbara. Email:
\texttt{cyx@ucsb.edu}}
}

\begin{document}

\maketitle
\thispagestyle{empty}
\pagestyle{empty}

\begin{abstract}

Animals such as squirrels and even bears have adapted to rapidly climb up trees and other complex vertical terrain, but achieving comparable agility has been a challenge for climbing robots. Existing robots often use walking gaits and move conservatively to stay in contact with the surface, which limits the range of dynamic maneuvers. In this paper we present a 290 g robot that, to our knowledge, is the first to climb vertically by bounding (with an aerial phase). We leverage a co-design workflow, in which the morphology and trajectory are jointly optimized for fast locomotion, subject to adhesion force limitations seen in spined grippers. The resulting trajectory includes a rapid maneuver that launches the robot vertically, and an aerial reorientation that brings the front grippers back to the surface using the rear leg as an inertial tail. We evaluate the resulting jump forces in 2D force space and demonstrate that the optimized morphology is capable of continuous climbing at a speed of 0.375 m/s (1.97 body lengths/s), and can also achieve ground locomotion and transition to a vertical surface. Our work proposes design insights for the jump-climbing maneuver and serves as an important step toward creating climbing robots with agility on par with that of animals.

%We show that the resulting prototype jumps up a vertical surface with an -\% aerial phase

\end{abstract}

\begin{comment}
\begin{IEEEkeywords}
   Mechanism Design of Mobile Robots, Multilegged Robots, Compliant joints and mechanisms
\end{IEEEkeywords}
\end{comment}

\section{Introduction}

Jumping and climbing allow robots to explore terrain that cannot be traversed by walking or rolling, and are especially useful for reaching locations that humans cannot, due to our size or for our safety. Jumping clears gaps and obstacles, while climbing provides sustained access to steep surfaces \cite{zhang2020jumpingreview, fang2023climbingreview}. Both behaviors use contact forces from the environment to support and accelerate the robot, avoiding the continuous lift required by aerial vehicles \cite{hsiao2025hybrid}. A combination of jumping and climbing could be promising for monitoring complex terrain such as arboreal ecosystems \cite{cannon2021extending}.

%For example, Hopcopter uses an elastic leg to bounce using ground contact, improving the endurance of an otherwise aerial platform \cite{bai2024hopcopter}. 

Many climbing robots demonstrate quasistatic walking gaits on vertical terrain. Initial hexapedal versions of RiSE were designed around pentapedal gaits, recirculating one leg at a time to maximize stability of their spined grippers \cite{spenko2008rise}.
The quadrupedal RiSE V3 used a combination of a crawl and bound to maintain at least two points of contact and could climb at a maximum speed of 0.21 m/s, or 0.3 body lengths/s \cite{haynes2009}. Waalbot, which adhered with fibrillar adhesives, used three-footed wheels in a differential drive configuration. The gecko-inspired Stickybot I and III used trot gaits, with diagonal limbs attaching and detaching together  \cite{stickybot1}\cite{stickybot3}. The 8 kg quadruped MARVEL used electropermanent magnetic feet to crawl, pace, and trot on ferromagnetic surfaces, achieving the fastest reported vertical walking speed for a climbing robot of 0.7 m/s, or 2.12 body lengths/s \cite{hong2022marvel}. 

There are also some examples of robots that climb dynamically. DynoClimber and ROCR translate a biological template based on cockroaches and geckos into robotic platforms that climb using a pendular motion parallel to the vertical surface, while keeping one foot in contact at all times \cite{goldman2006template,lynch2012dynoclimber,provancher2010rocr}. ParkourBot achieved stable climbing gaits bouncing between two opposing walls while supported by an air bearing table angled 9º from horizontal \cite{parkourbot}. 
Trident, a 2-joint lamprey-inspired climbing robot jumps up surfaces 70º from horizontal using dynamic tail movements \cite{vanstratum2023lamprey}.

Lin et al. demonstrated a tethered robot that moves up a vertical surface using a spring-driven jump mechanism and a weighted tail to reorient the body toward the vertical surface \cite{lin2017walljumping}. Their system established the feasibility of repeated wall jumps, but required a custom array of wedges on the wall surface, a large tail mass, and a tether for power. Further, its jumps were limited to 2.5 cm approximately every 6.2 s (0.004 m/s). 
%This leaves open how the morphology and motion of a jump-climbing robot should be designed together for faster untethered climbing.
Finally, a previous quadrupedal robot is capable of jumping and perching onto tree bark, but left climbing as future work \cite{xu2025pinto}.

Accordingly, to the best of our knowledge, there is no previous untethered climbing robot that can scale a vertical wall using an aerial phase.
Yet, climbing with an aerial phase opens the door for dynamic maneuvers that are desirable for navigating through complex terrain full of obstacles and gaps between footholds. 
%Animal locomotion strategies suggest that sustained attachment is not always necessary. 
For instance, squirrels use an aerial phase to bound rapidly up trunks and along or between branches, and sometimes use brief contacts with vertical surfaces to redirect their motion during a leap \cite{hunt2021}. Despite their large size, bears have also been recorded to climb with a bounding gait that includes an aerial phase at high speeds \cite{bearclimbingYT}. 
%Animal climbing gaits often contain both contact and aerial phases, motivating a robot that gains height through repeated jumps rather than a quasistatic gait. 
Robots that have comparable agility leaping and bounding in trees remains an open challenge.

\begin{figure}[t]

    \centering
    \includegraphics[width=\columnwidth]{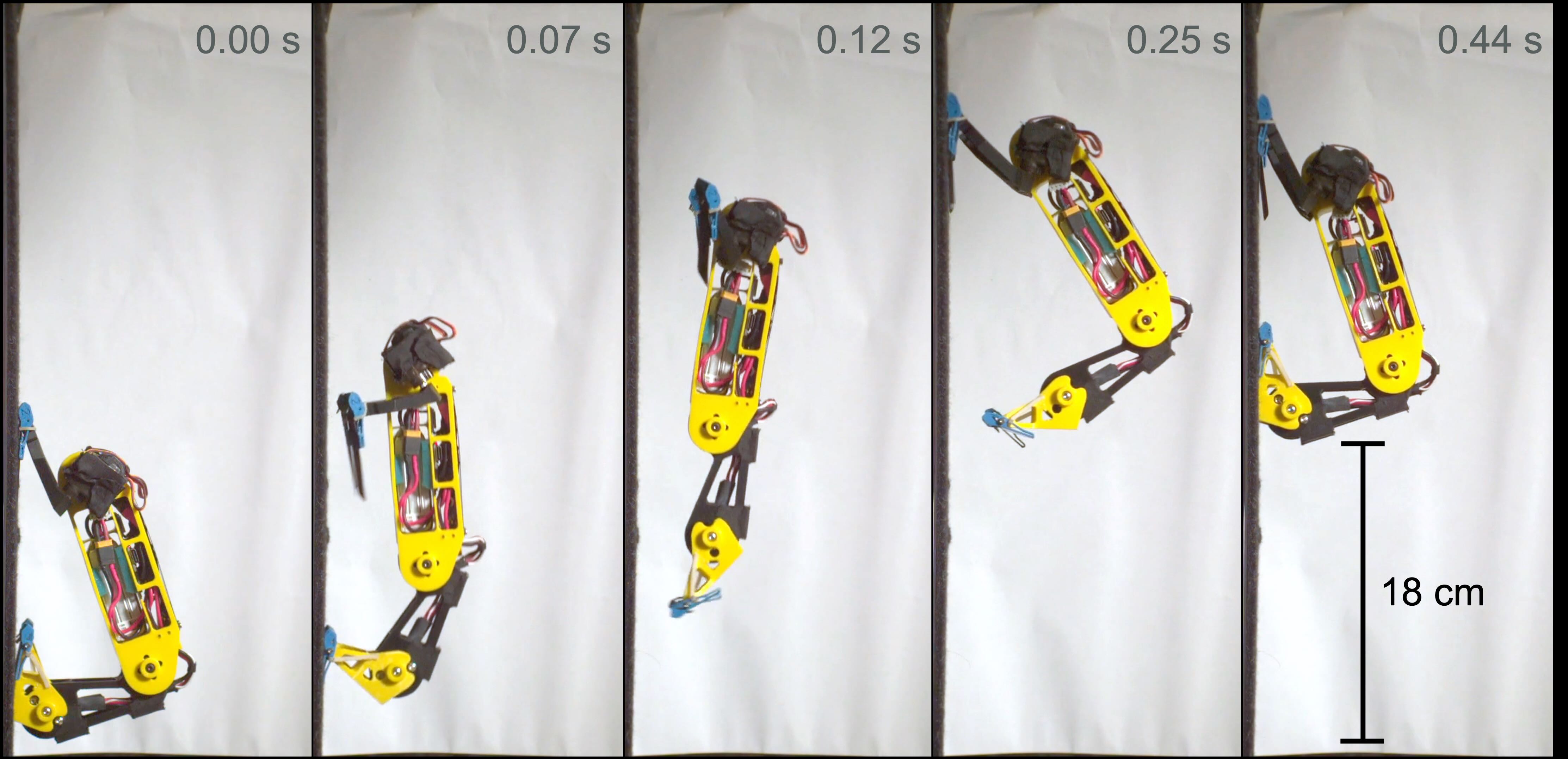}
    \caption{The climbing robot jumps up a flat vertical surface, covering nearly a body length per stride. A fully aerial phase is seen in the third panel. In the fourth panel, it swings its rear legs towards its torso to bring its front grippers back to the wall. Video: \href{https://qwertpas.github.io/jump-climb/}{qwertpas.github.io/jump-climb}}
    \label{fig:fivejumppanels3_2}
    \vspace{-1.25em}
\end{figure}

In this work, we present a vertical jump-climbing robot that repeatedly launches from and reattaches to a carpeted wall with spined grippers. We propose a three-stage co-design process that jointly searches the robot morphology and jump trajectory design spaces, then refines the resulting design in hardware. 
We characterize the resulting locomotion pattern in 2D force space and compare to the planned reference. We demonstrate the robot climbing both flat and cylindrical walls, achieving a speed of 1.97 bodylengths/s, faster than previous spined climbers by body length on a vertical surface \cite{lynch2009highspeeddynoclimber,birkmeyer2011clash}.
Finally, we found that the robot can use the same limb mechanisms designed for climbing to perform ground jumping and perching, advancing the broader goal of a squirrel-like robot that can move between the ground and vertical terrain. 
%Rather than add a mechanism used only for climbing, we investigate jump-climbing using the same legs and grippers used for ground locomotion, jumping, and perching.

%Ask AI for more citations!

\begin{figure}
    \centering
    \includegraphics[width=\columnwidth]{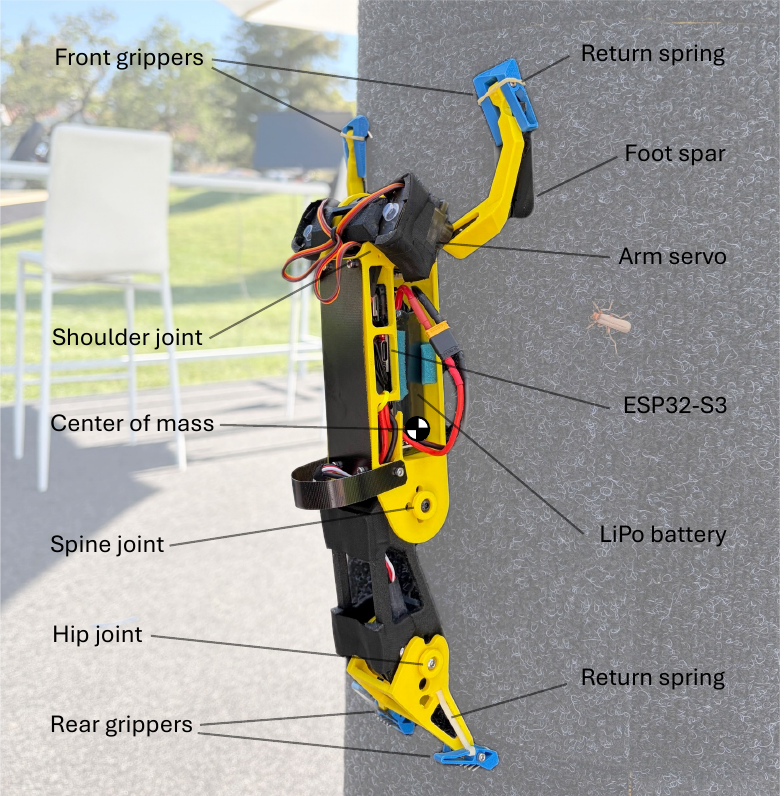}
    \caption{The robot is perched on a 12-inch diameter carpeted tube. The center of mass is slightly above the spine joint.}
    \label{fig:beetle}
    \vspace{-1.25em}
\end{figure}

\section{Design}

\begin{figure}
    \centering
    \includegraphics[width=\columnwidth]{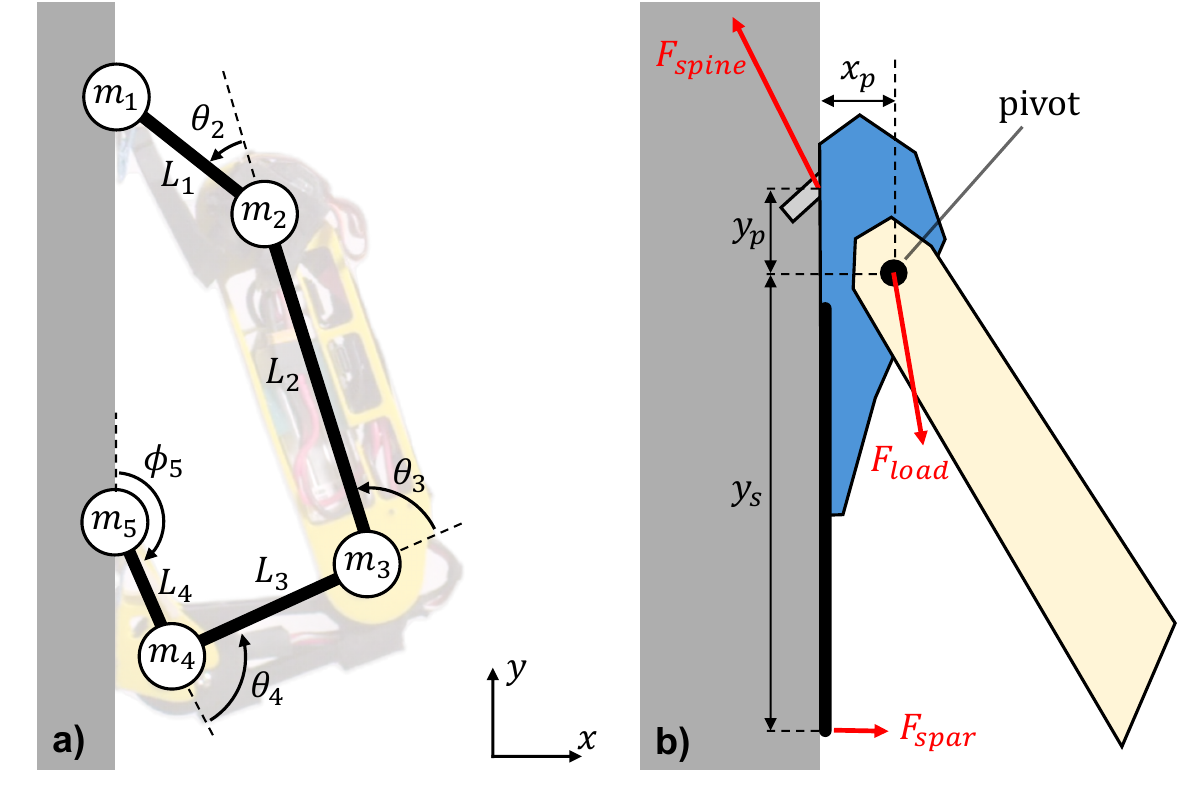}
    \caption{a) Planar 4-link model of the robot. $\theta_2$, $\theta_3$, and $\theta_4$ are actuated. $m_5$ is treated as the floating base and constraint forces are applied to $m_1$ and $m_5$ to simulate wall contact depending on the phase of the jump. b) Forces and moments acting on the gripper spines, pivot, and spar are balanced when the robot is adhered to the surface. }
    \label{fig:schematic}
    \vspace{-1.25em}
\end{figure}

An overview of the robot design is shown in Fig. \ref{fig:beetle}. Loosely inspired by bounding gaits in squirrels, the system has a symmetric 4-link, 3-joint design in the sagittal plane (Fig. \ref{fig:schematic}a), driven by brushless motors. Spined grippers are pinned at the end of two front arms and at the end of the single rear leg. To aid gripper detachment and to accommodate both flat and cylindrical vertical surfaces, a low power servo adjusts the abduction/adduction of each front arm.

\subsection{Gripper design and characterization} \label{sec:gripper}

To stay attached to a vertical surface, climbing robots generally apply adhesion forces with methods based on negative air pressure, electromagnetic and electrostatic interactions, chemical and dry adhesives, and mechanical spines \cite{chattopadhyay2018adhesion}. Spined grippers are relatively simple to produce, can adhere to a wide range of natural materials, penetrating soft substrate like plants or tree barks or interlock with small asperities on a rough, hard surfaces, like rock \cite{haynes2009,Microspine,stickybot3}. Because the focus of this paper is the design of the robot rather than the spines, we use an easy-to-grip surface---3 mm thick polyester carpet---and simple spines---2.54 mm pitch male pin headers. 

%These mechanisms can support reliable attachment, but many are specialized for climbing and add hardware that may be cumbersome during ground locomotion.

For a basic spined gripper, we can write down a static equilibrium model that helps inform design. We consider the gripper force to be the sum of the hooking force at the spines and the normal force at the spar, a protrusion at the back of the foot. The resulting total  force applied by the wall contact to the gripper is $f=F_{spine}+F_{spar}$, which opposes the load of the robot $F_{load}$ (Fig. \ref{fig:schematic}b). Horizontal and vertical force balances result in the expressions \(F_{spine,x} = F_{load,x} + F_{spar}\) and \(F_{load,y} = F_{spine,y} \), where the subscript $x$ and $y$ represents forces normal and tangential to the surface. We can solve for $F_{spar}$ with a moment balance about the gripper pivot (assumed to apply no frictional torques), and write the adhesion force at the spines:

%moment balance: \(F_{spine,x}y_p - F_{spine,y}x_p + F_{spar}y_s = 0\). 
 % \[F_{spine,x}=F_{load,x}+F_{spar}\]
 % \[F_{spar}=\frac{F_{spine,y}x_p-F_{spine,x}y_p}{y_s}\]
 % \[F_{spine,x}=F_{load,x}+\frac{F_{spine,y}x_p-F_{spine,x}y_p}{y_s}\]
 % \[F_{spine,x}y_s=F_{load,x}y_s+F_{spine,y}x_p-F_{spine,x}y_p\]
 % \[F_{spine,x}(y_s+y_p)=F_{load,x}y_s+F_{spine,y}x_p\]
 % \[F_{spine,x}=\frac{F_{load,x}y_s+F_{spine,y}x_p}{y_s+y_p}\]
 % \[F_{spine,x}=\frac{F_{load,x}y_s+F_{load,y}x_p}{y_s+y_p}\]
%rewrite to separate F_spar:
\[
F_{\mathrm{spine},x} = F_{\mathrm{load},x}+\underbrace{
    \frac{F_{\mathrm{load},y}x_p-F_{\mathrm{load},x}y_p}{y_s+y_p}
}_{F_{\mathrm{spar}}}\]

Thus, decreasing $x_p$ (moving the pivot closer to the wall) should reduce adhesion force necessary at the spines. Additionally, increasing $y_p$ (moving the pivot downwards) and increasing $y_s$ (longer protrusion) are also beneficial, up until $F_{spar}$ goes to zero. A physical interpretation is given in \cite{stickybot3}: increasing the length of the spar reduces the normal force necessary to counteract the pitchback moment that is created by the downwards load offset from the wall.

\begin{figure}
    \centering
    \includegraphics[width=\columnwidth]{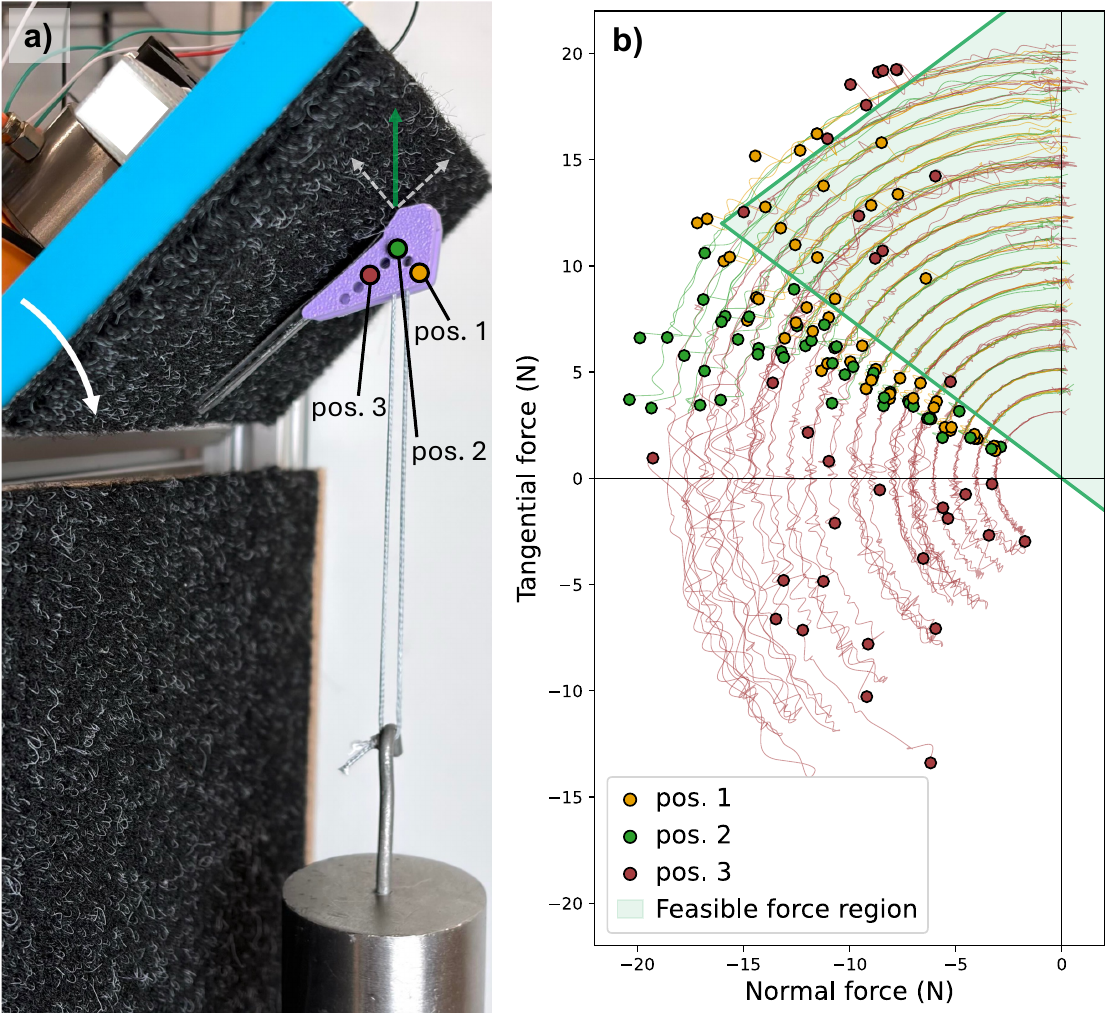}
    \caption{a) Test setup for measuring adhesion limits of gripper.  b) Trials begin on the vertical axis (pure tangential force) and as the surface tilts, the gripper force relative to the surface gradually sweeps counterclockwise until the gripper falls off, at which point a colored dot is marked on the plot. The force trajectory is plotted with thin traces so the radius of each arc depicts the test weight. Position 2 (green) generally achieves higher adhesion force than Position 1 (orange) and always requires upwards tangential force to stay adhered. Position 3 (red) stays adhered even with downwards tangential force and sometimes does not fall at all, so no colored dot is shown. A conservative feasible region is estimated for the chosen Position 2 using two linear inequalities (shaded green).}
    \label{fig:region}
    \vspace{-1.25em}
\end{figure}

We experimentally evaluated the effect of the wrist pivot position on the space of 2D forces $f$ achievable by the gripper by hanging weights from a test gripper with multiple possible pivot locations, then slowly rotating the surface to apply forces at varying angles (Fig. \ref{fig:region}). 
Three pivot positions were tested (positions 1, 2, 3), with weights ranging from 400 g to 2000 g and angles ranging from vertical to approximately 135º from vertical.

As predicted, we find that a pivot farther from the wall (pos. 1) achieved less negative normal force, or adhesion, $F_{load,x}$. We also found that too low of a pivot (pos. 3) resulted in inconsistent adhesion and variation across trials, occasionally holding the loads at extreme angles without falling off. This is due to the spar losing contact with the surface ($F_{spar} = 0$), causing the spines to rotate into the carpet substrate and getting stuck after hooking onto loops deep inside the carpet, which would be undesirable for rapid repeated jumps. We choose a pivot position as close to the wall as possible under packaging constraints and high enough to reliably release (pos. 2). 

Microspine grippers designs on rough rigid surfaces often use a limit surface to quantify safe forces, which describes an adhesion force limit that increases as the force becomes more tangential to the surface, and a maximum force limit dependent on spine and asperity strength \cite{asbeckdesigning}. Similarly, we observe consistent adhesion in a sector within some angle from the vertical surface for the chosen pivot position. We approximate the safe region using two linear inequalities to provide simple convex contact force constraints for the trajectory optimization process in Section \ref{sec:trajopt}.

% We define the feasible contact-force region as
% \[
% \mathcal{F}
% =
% \left\{
% \mathbf{f}=
% \begin{bmatrix} f_x \\ f_y \end{bmatrix}
% \in\mathbb{R}^2
% \;\middle|\;
% -\frac{3}{4}f_n \leq f_t
% \leq \frac{3}{4}f_n + 24\,\mathrm{N}
% \right\},
% \]
% where $f_x$ and $f_y$ denote the normal and tangential
% force components, respectively, using the sign convention
% shown in Fig.~\ref{fig:region}.

\subsection{Co-design of trajectory and morphology}\label{sec:codesign}

For trajectory design (without co-design of morphology), climbing robots have used optimization-based motion planning to respect adhesion force limits, including quasi-static planning for dry adhesives \cite{boscariol2013gait} and planning with spined gripper limit surfaces \cite{shirai2022climbing}. For leaping robots, trajectory optimization additionally considers rigid body dynamics for takeoff, flight, and landing, enabling high jumps and aerial maneuvers on the MIT Cheetah \cite{nguyen2019jumping}. 

At the same time, morphology design affects robot capabilities and may change the optimal trajectory. Mass distribution influences inertial reorientation \cite{libby2016reorientation}, and body dimensions affect adhesion forces required for climbing \cite{ahmed2015design}. 

Co-design methods can address both design challenges by jointly optimizing the motion and physical parameters \cite{spielberg2017codesign,ha2018codesign}. In climbing robot design in simulation, Awan uses a co-design procedure alternating controller weight tuning and morphology search of a two-legged wall climbing robot to minimize energy consumption. They use a spring-damper contact model and allow the control policy to vary the adhesion force \cite{awan2026}. Trident jointly optimizes its link lengths and actuation trajectory for jumping from a single gripper, but reattachment is not considered due to being on a 70º slope \cite{vanstratum2023lamprey}. To our knowledge, there is gap in the current literature for a co-design procedure of trajectory and morphology for completely vertical climbing that is verified on hardware. 
%with spined grippers.

In our approach, we use a nested approach similar to Fadini et al. \cite{fadini2024codesign}, in which an outer loop searches morphology parameters and an inner loop optimizes the trajectory for each candidate, allowing each morphology to be evaluated with a motion adapted to its dynamics. In this work, we hypothesize that a 4-link kinematic structure is sufficient to achieve vertical bounding. We fix the kinematic structure, gripper force constraints, and actuator selection, then optimize link lengths, mass distribution, and joint trajectories. 

% jump-climbing motion is very influenced by body dynamics, mass distribution
% Due to difficult-to-specify practical constraints and the reliance of experimental data, the gripper design and force constraints are fixed throughout the codesign optimization process.

\begin{figure*}
    \centering
    \resizebox{0.96\textwidth}{!}{\definecolor{workflowblue}{HTML}{D7E9F2}
\definecolor{workflowsand}{HTML}{F3E5C3}
\definecolor{workflowline}{HTML}{66727A}

\begin{tikzpicture}[
    x=1cm,
    y=1cm,
    >={Stealth[length=4pt,width=3pt]},
    flow/.style={draw=workflowline, line width=0.45pt, -{Stealth[length=4pt,width=3pt]}},
    box/.style={
        draw=workflowline,
        rounded corners=3pt,
        line width=0.45pt,
        text width=3.05cm,
        minimum height=0.82cm,
        align=center,
        inner xsep=3pt,
        inner ysep=4pt,
        font=\fontfamily{cmss}\fontsize{10}{12}\selectfont
    },
    model/.style={box, fill=workflowblue},
    task/.style={box, fill=workflowsand},
    note/.style={
        font=\fontfamily{cmss}\fontsize{11}{12.5}\selectfont,
        text=black!75,
        align=center,
        fill=white,
        inner xsep=2pt,
        inner ysep=1pt
    },
    stage/.style={font=\fontfamily{cmss}\fontsize{11}{12.5}\selectfont, anchor=north west, align=left, inner xsep=0pt}
]

% stage 1
\node[task]  (initial-shape) at (2.50,5.08) {initial morphology};
\node[model] (initial-model) at (6.40,5.08)
    {symbolic dynamics\\model};
\node[task]  (initial-goals) at (2.50,2.78) {initial constraints and\\objectives};
\node[model] (low-res) at (6.40,2.78) {low-res traj. opt.};
\node[task]  (human-eval) at (2.50,0.48)
    {\mbox{visual evaluation}};
\node[model] (high-res-one) at (6.40,0.48) {high-res traj. opt.};

\draw[flow] (initial-shape) -- (initial-model);
\draw[flow] (initial-goals) -- (low-res);
\draw[flow] (initial-model) -- (low-res);
\draw[flow] (low-res) -- (high-res-one);
\draw[flow] (high-res-one) -- (human-eval);
\draw[flow] ([yshift=0.23cm]low-res.east) -- ++(0.90,0)
    -- ++(0,-0.46) -- ([yshift=-0.23cm]low-res.east);
\draw[flow] (human-eval.west) -- ++(-1.05,0) |- (initial-shape.west);
\draw[flow] ([xshift=-1.05cm]human-eval.west) |- (initial-goals.west);

% stage 2
\node[task]  (shape-limits) at (10.90,5.08) {morphology\\constraints};
\node[model] (sampler) at (14.80,5.08) {CMA-ES sampler};
\node[model] (search-model) at (14.80,2.78)
    {symbolic dynamics\\model};
\node[model] (high-res-two) at (14.80,0.48) {high-res traj. opt.};

\draw[flow] (shape-limits) -- (sampler);
\draw[flow] (sampler) -- (search-model);
\draw[flow] (search-model) -- (high-res-two);
\draw[flow] (high-res-one) -- (high-res-two);
\coordinate (feedback-spine) at (17.38,0.66);
\draw[flow] ([yshift=0.18cm]high-res-two.east)
    -- (feedback-spine) |- (sampler.east);

% stage 3
\node[task]  (mechanical) at (23.00,5.08) {mechanical design\\with optimized\\dimensions};
\node[model] (reoptimize) at (23.00,2.78) {trajectory\\reoptimization};
\node[task]  (hardware) at (23.00,0.48) {tuning on hardware};

\coordinate (realization-spine) at (19.50,0.30);
\draw[flow] ([yshift=-0.18cm]high-res-two.east)
    -- (realization-spine) |- (mechanical.west);
\draw[flow] (realization-spine |- reoptimize.west) -- (reoptimize.west);
\draw[flow] (mechanical) -- (reoptimize);
\draw[flow] (reoptimize) -- (hardware);

% Opaque labels are drawn last so they sit in front of every connector.
\node[note] at (6.40,3.93)
    {$\mathbf{M}(\mathbf q),\,\mathbf{h}(\mathbf q,\dot{\mathbf q}),\,\mathbf{J}(\mathbf q)$};
\node[note] at (6.40,1.63) {best candidates};
\node[note, anchor=west, align=left] at (8.95,2.78) {iteration on\\initial guess};
\node[note] at (14.80,3.93) {morphology\\parameters};
\node[note] at (14.80,1.63)
    {$\mathbf{M}(\mathbf q),\,\mathbf{h}(\mathbf q,\dot{\mathbf q}),\,\mathbf{J}(\mathbf q)$};
\node[note] at (10.60,0.48) {seed trajectory};
\node[note] at (17.35,2.78) {fitness\\score};
\node[note] at (19.50,1.63) {best morphology/\\trajectory pair};
\node[note] at (23.00,3.83) {URDF};
\node[note] at (23.00,1.63) {joint trajectories};

% Align stage titles with the left edges of their yellow-box columns.
\node[stage] at (initial-shape.west |- 0,6.85)
    {Stage 1: Exploration with\\single morphology};
\node[stage] at (shape-limits.west |- 0,6.85)
    {Stage 2: Trajectory/Morphology\\codesign loop};
\node[stage] at (mechanical.west |- 0,6.85)
    {Stage 3: Realization\\and refinement};

\end{tikzpicture}}
    \caption{Co-design workflow. Yellow boxes denote steps where the human is involved, and blue boxes are automated.}
    \label{fig:codesign-workflow}
    \vspace{-1.25em}
\end{figure*}
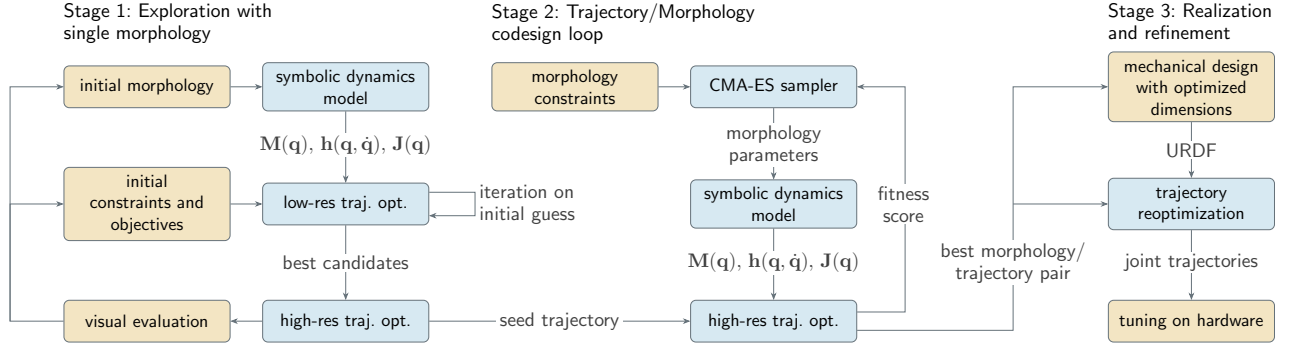

\subsection{Dynamics}\label{sec:dynamics}

For simplicity, the robot is modeled in the sagittal plane as 5 point masses connected by 4 rigid links, with the rear foot as a floating base (Fig. \ref{fig:schematic}a). The configuration is represented by the 6-dimensional vector \(q = [x_5, y_5, \phi_5, \theta_4, \theta_3, \theta_2]\).

We express the forward kinematics, kinetic energy
$T(\mathbf{q},\dot{\mathbf{q}})$, potential energy $V(\mathbf{q})$,
and Lagrangian $\mathcal{L}=T-V$ symbolically in CasADi, then use
automatic differentiation \cite{casadi} to construct the mass
matrix $\mathbf{M}(\mathbf{q})$, bias force vector
$\mathbf{h}(\mathbf{q},\dot{\mathbf{q}})$, and contact Jacobians
$\mathbf{J}_1(\mathbf{q})$ and $\mathbf{J}_5(\mathbf{q})$ at the
front and rear gripper positions $\mathbf{p}_1$ and $\mathbf{p}_5$:
\[
\mathbf{M}(\mathbf{q})
= \frac{\partial^2 T}{\partial \dot{\mathbf{q}}^2},
\]
\[
\mathbf{h}(\mathbf{q},\dot{\mathbf{q}})
= \frac{\partial}{\partial \mathbf{q}}
  \left(\frac{\partial \mathcal{L}}{\partial \dot{\mathbf{q}}}\right)
  \dot{\mathbf{q}}
  - \frac{\partial \mathcal{L}}{\partial \mathbf{q}},
\]
\[
\mathbf{J}_1(\mathbf{q})
= \frac{\partial \mathbf{p}_1(\mathbf{q})}{\partial \mathbf{q}},
\qquad
\mathbf{J}_5(\mathbf{q})
= \frac{\partial \mathbf{p}_5(\mathbf{q})}{\partial \mathbf{q}}.
\]

The dynamics are then expressed by the manipulator equation:
\[
\mathbf{M}(\mathbf{q})\ddot{\mathbf{q}}
+ \mathbf{h}(\mathbf{q},\dot{\mathbf{q}})
= \mathbf{B}\boldsymbol{\tau}
+ \sum_{i \in \mathcal{C}}
  \mathbf{J}_i(\mathbf{q})^\top \boldsymbol{f}_i
\]
where $\boldsymbol{\tau}=[\tau_4,\tau_3,\tau_2]^\top$ are the actuator torques,
$\mathbf{B}$ maps these torques directly to the joints $\theta_4$, $\theta_3$, and $\theta_2$,
$\mathcal{C}$ is the set of active gripper contacts, and
$\boldsymbol{f}_{i}$ is the cartesian contact force
acting on the robot at gripper $i$.

\subsection{Trajectory optimization}\label{sec:trajopt}

Subject to these dynamics, we formulate a trajectory optimization problem with direct collocation for one jump, attempting to maximize jump speed. To avoid requiring the optimizer to reason about discrete contact switching, we prescribe a fixed contact schedule based on a bounding gait:
\begin{itemize}
    \item Phase 1 ($\mathcal{C}=\{1,5\}$): Both front and rear
    grippers contact the surface, forming a closed chain.
    \item Phase 2 ($\mathcal{C}=\{5\}$): Only the rear gripper in contact.
    \item Phase 3 ($\mathcal{C}=\emptyset$): The robot is
    completely airborne.
    \item Phase 4 ($\mathcal{C}=\{1\}$): Only the front gripper in contact.
    \item Phase 5 ($\mathcal{C}=\{1,5\}$): Both grippers in contact again.
\end{itemize}
% Each phase is represented by a panel in Fig. \ref{fig:fivejumppanels3_2}. 

Each phase is allocated a fixed number of nodes (\(N_1,N_2,N_3,N_4,N_5\)) at which the optimizer gets to decide states $q_{k}$ and $\dot{q}_{k}$, actuator torques $\tau_k$, and contact forces $f_{i,k}$, and the dynamics are evaluated at each node $k$. The optimizer decides a different time step between nodes for each phase ($\Delta t_1$,$\Delta t_2$,$\Delta t_3$,$\Delta t_4$,$\Delta t_5$) to control how long each phase of the jump lasts. The optimization objective is then to maximize the final height gain divided by the total jump period: 
$H / \sum_{p=1}^{5}N_p \Delta t_p$.
% $H / (N_1 \Delta t_1+N_2 \Delta t_2+N_3 \Delta t_3+N_4 \Delta t_4+N_5 \Delta t_5)$

Each node adds a dynamics constraint (Section \ref{sec:dynamics}) according to trapezoidal discretization. At each node, the contact force at each gripper on the wall $f_i$ is constrained to be within the safe region characterized in Section \ref{sec:gripper}. 

% \[\boldsymbol{f}_{i,k}\in\mathcal{F},
% \qquad i\in\mathcal{C}\]
% \[\mathbf q_{k+1}
% =\mathbf q_k+\frac{\Delta t}{2}
% (\dot{\mathbf q}_k+\dot{\mathbf q}_{k+1})\]
% \[\dot{\mathbf q}_{k+1}
% =\dot{\mathbf q}_k+\frac{\Delta t}{2}
% (\ddot{\mathbf q}_k+\ddot{\mathbf q}_{k+1})\]
When the gripper $i$ attaches to the wall,  generalized velocity changes from $\dot{\mathbf{q}}^-$ to $\dot{\mathbf{q}}^+$ such that the front gripper stops. This is enforced using the constraint \(\mathbf J_i\dot{\mathbf q}^{+}=\mathbf0\), and the impulse $\boldsymbol{\Lambda}_i$ at the gripper changes the generalized momentum:
\({\mathbf M(\dot{\mathbf q}^{+}-\dot{\mathbf q}^{-})
=\mathbf J_i^\top\boldsymbol{\Lambda}_i}\).
2D gripper attachment impulses $\boldsymbol{\Lambda}_1$ and $\boldsymbol{\Lambda}_5$ exist at the phase 3-4 and phase 4-5 transitions, respectively, and are decision variables for the trajectory optimization.

To avoid contacting the wall during the aerial phase, each foot not pinned ($i \notin C$) satisfies $p_{i,x} \geq r_{foot}$, where $r_{foot}$ represents the minimum foot clearance. Similarly, the horizontal position of each inner joint ($p_{2,x},p_{3,x},p_{4,x}$) is constrained to be at least a distance one motor radius away from the wall. For continuous jumps, the last state of phase 5 is constrained to equal the first state of phase 1, except for height. Finally, we enforce box constraints on the timestep, motor torques, and joint angles.

%The decision variables consist of the states $q_{k}$  and $\dot{q}_{k}$ , actuator torques $\tau_k$, contact forces $f_{i,k}$, contact impulses $\boldsymbol{\Lambda}_1$ and $\boldsymbol{\Lambda}_5$, and time step $\Delta t$ for each for the five phases.

% \sum_{s \in \{1,2,3,4,5\}}
% N_1\Delta t_1+N_2\Delta t_2+N_3\Delta t_3+N_4\Delta t_4+N_5\Delta t_5

% In summary, the trajectory optimization solves:
% \[
% \begin{aligned}
% \underset{\substack{
%  \mathbf{q},\,\dot{\mathbf{q}},\,\boldsymbol{\tau},\,
%  \boldsymbol{\lambda},\,\boldsymbol{\Lambda}, \Delta t
% }}{\operatorname{minimize}}
% \quad&
% \frac{-H}{\sum_{s}N_s \Delta t_s}
% \\[2pt]
% \text{subject to}\quad&
% \mathbf{q}_{s,k+1}
% =\mathbf{q}_{s,k}
% +\frac{\Delta t_s}{2}
%  (\dot{\mathbf{q}}_{s,k}+\dot{\mathbf{q}}_{s,k+1}),
% \\
% &
% \dot{\mathbf{q}}_{s,k+1}
% =\dot{\mathbf{q}}_{s,k}
% +\frac{\Delta t_s}{2}
%  (\ddot{\mathbf{q}}_{s,k}^{-}+\ddot{\mathbf{q}}_{s,k}^{+}),
% \\
% &
% \mathbf{J}_i(\mathbf{q}_{s,k})\dot{\mathbf{q}}_{s,k}=0,
% \qquad i\in\mathcal{C}_s,
% \\
% &
% \boldsymbol{f}_{i,s,k}\in\mathcal{F},
% \qquad i\in\mathcal{C}_s,
% \\
% &
% -\tau_{\max}
% \leq\boldsymbol{\tau}_{s,k}\leq\tau_{\max},
% \\
% &
% \theta_{\min}
% \leq\boldsymbol{\theta}_{s,k}\leq\theta_{\max},
% \\
% &
% \Delta t_s^{\min}\leq\Delta t_s\leq\Delta t_s^{\max},
% \end{aligned}
% \]
% with $s\in\{1,\ldots,5\}$ to index the contact phases and
% $k\in\{0,\ldots,N_s-1\}$ to index each node within each phase.

The nonlinear optimization problem is solved using IPOPT in CasADi \cite{casadi}. We find that a high node count often fails to converge depending on the random initial guess. Therefore, we use a coarse resolution \((N_1,N_2,N_3,N_4,N_5)=(18,9,15,9,12)\) for initial exploration from a randomized seed. They are then refined to higher resolution \((27,14,23,14,18)\) using linear interpolation of each decision variable for each node as an initial guess. For Stage 1 of our co-design workflow (Fig. \ref{fig:codesign-workflow}), we visualize the motion and manually iterate on the trajectory optimization setup until the intended motion is produced.

%If we consider link $L_2$ to be the body, then the front arm is $L_1$ and the rear leg consists of $L_3$ and $L_4$. 

\subsection{Geometry optimization}

Once we have a procedure to optimize trajectories for a given morphology, we then use a CMA-ES \cite{hansen2016cmaes} outer loop to optimize over the morphology design space, which consists of four link lengths and three inner masses (Stage 2 in Fig. \ref{fig:codesign-workflow}). For each generation, each candidate morphology has a trajectory optimized for it, and the resulting climbing speed is used as the fitness objective to update the next generation. We chose this sampling-based search because it does not require gradients from trajectory optimization.

Each of the four link lengths may be varied between a minimum of 40 mm, which represents motor packaging constraints, and a maximum of 120 mm, which was chosen to avoid an excessively long robot with wasted structural mass. The distal point masses $m_1$ and $m_5$ were both fixed at 10 grams, while the mass distribution across the three inner masses are allowed to vary. $m_2$, $m_3$, and $m_4$ were each constrained to a minimum of 40 g, the mass of the motor, and their sum is constrained to be 250 g. As an initial guess for CMA-ES, we set each link to the middle of the range, at 80 mm each, and distribute the masses evenly, at 83 g each.

\begin{figure}
    \centering
    \includegraphics[width=\columnwidth]{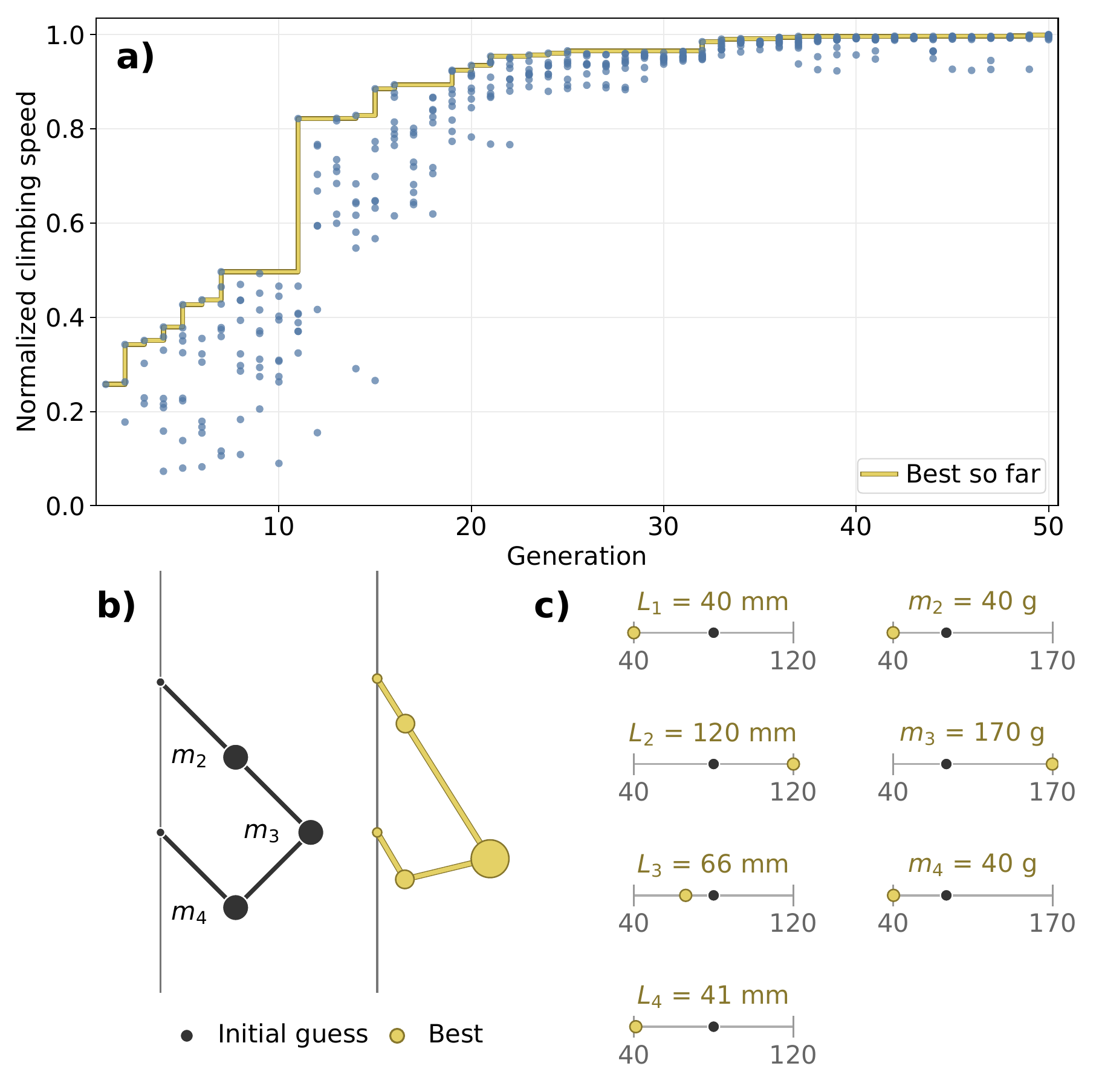}
    \vspace{-2em}
    \caption{a) Over 50 CMA-ES generations, the best morphology/trajectory combination improves over the initial generation by approximately 3x, with progress faster in earlier generations. b) The initial morphology is sketched in gray, and outcome of the CMA-ES optimization in yellow. c) The resulting morphology has short end links, a long $L_2$, and concentrates mass at the spine joint.}
    \label{fig:cmaes}
    \vspace{-1.25em}
\end{figure}

Each generation, a population of 9 candidates is evaluated using 9 parallel CPU worker processes. On a Linux virtual machine with an AMD EPYC 7J13 processor, the co-design loop completed 50 generations in 188 minutes. Best climbing speed improved until the population converged to a local maximum; the last generation has length variation of less than 1 mm and mass variation of less than 1 g. The resulting morphology has a three notable characteristics (Fig. \ref{fig:cmaes}): 
\begin{itemize}
    \item mass concentrated at the spine joint ($m_3>m_2,m_4$)
    \item short distal links ($L_1,L_4$) 
    \item front two links combined are longer than rear two links combined ($L_1+L_2>L_3+L_4$). 
\end{itemize}

\begin{figure}[t]
    \centering
    \includegraphics[width=0.95\columnwidth]{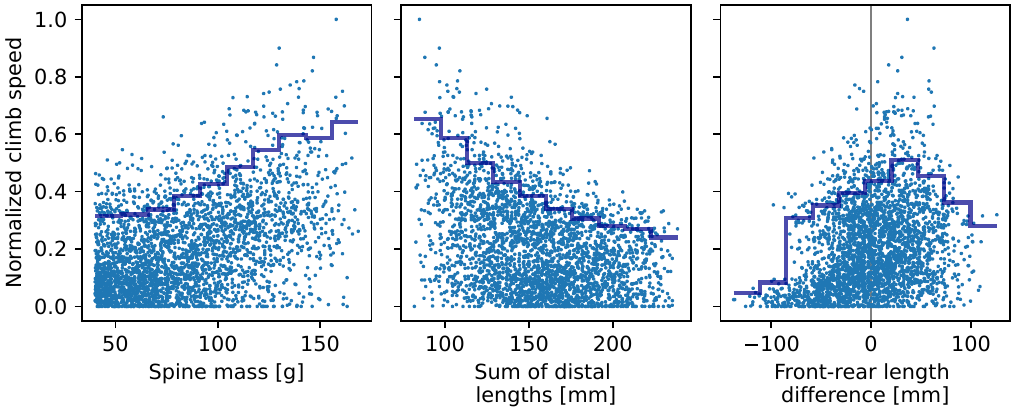}
    \vspace{-0.5em}
    \caption{Across 3031 random morphology samples, higher spine masses ($m_3$), shorter distal links ($L_1+L_4$), and longer front vs rear lengths ($L_1+L_2-L_3-L_4$) produced more higher performing trajectories. Each morphology sample is plotted in light blue. The 95\% percentile speed is plotted in dark blue, split over 10 bins for each parameter range.}
    \label{fig:sobol}
    \vspace{-1.25em}
\end{figure}

To probe broader trends over the parameter space, we sampled morphologies with a uniform coverage of feasible link lengths and masses using a Sobol sequence and optimized a coarse resolution trajectory for each. Randomized initial trajectory guesses mostly resulted in solver infeasibility across the wide range of morphologies, so the same seed trajectory selected for the initial generation of the CMA-ES run was used as an initial guess for each trajectory optimization. The resulting climbing speed for each randomly sampled morphology is plotted against three characteristics in Fig. \ref{fig:sobol}. The random sampling process takes over 8 hours and cannot indicate a clear optimum, but we find that the trends agree with the best result of the CMA-ES run: higher spine mass, shorter distal links, and a bias towards longer front links are more likely to achieve higher climbing speeds.

\subsection{Hardware realization}

We use the optimized geometry and mass distribution and then adjust the design for fabrication constraints. Both the hardware and control use this process of optimization followed by manual tuning (Fig. \ref{fig:codesign-workflow}). For example, to concentrate mass at the spine joint, the heaviest component, a 46 g battery, is placed close to the spine motor while the other links are hollowed out to reduce mass.

The robot must survive falls from several body lengths up the wall yet be rigid enough for precise control of position relative to the surface. We balance rigidity and robustness using a combination of FDM 3D printed materials: critical joints are stiff carbon fiber-reinforced polyphthalamide nylon (BambuLab PPA-CF, black), while slender protrusions and electronics enclosures are impact-resistant 68D thermoplastic polyurethane (BambuLab TPU for AMS, yellow). Sagittal joints are actuated by backdrivable motors and ball bearings support both sides to prevent radial loads on the motor.

An ESP32-S3 module serves as the central robot controller to command motors to follow reference trajectories without additional sensor feedback. A laptop communicates wirelessly with the ESP32 to load trajectory references, start trajectory runs, and download logged telemetry. A 3S LiPo battery powers the robot, chosen for high current output at the desired mass. For each sagittal joint (shoulder, spine, and hip), we use a direct-drive brushless motor module for its rapid response and high power density when paired with the high current battery. Each motor module contains an integrated encoder and electronic speed controller (ESC) that runs a 1000 Hz proportional-derivative control loop to track the commanded joint angle. The ESP32 sends each ESC a commanded joint angle and a voltage feedforward (proportional to desired joint torque) at 500 Hz, then logs joint angle telemetry from each ESC at 167 Hz. Arm abduction/adduction are actuated by MG90S servos, which receive PWM angle commands at 250 Hz.

\begin{table}[b]
    \vspace{-1.5em}
    \centering
    \caption{Robot properties and components}
    \begin{tabular}{|c|c|} \hline 
         Mass& 290 g\\ \hline 
         Idle length& 19 cm \\ \hline 
         Battery& Ovonic 3S 550mAh 130C LiPo \\ \hline 
         Shoulder/spine/hip & 3x Vertiq 23-06 2200KV BLDC \\ \hline
         Arm actuator & 2x MG90S servo\\ \hline 
         Structure& 3D printed TPU, PPA-CF \\ \hline
         Microcontroller& ESP32-S3 Supermini \\ \hline
         Estimated BOM cost& 460 USD \\ \hline

        % Cost estimated:
        % - 6 for battery
        % - 391 for 3x BLDC
        % - 8 for 2x servo
        % - 5 for microcontroller
        % - 7 for TPU
        % - 30 for PPA
        % - 15 for wires, screws, bearings
        
    \end{tabular}
    \label{tab:table}
\end{table}

%Bearings support the joints on both sides. The front arms must be stiff enough to maintain gripper position under load, and the arm servos are placed to clear the tree throughout the range of motion. We arrange the components to keep the center of mass close to the optimized result. 

Each gripper passively rotates about the pivot location identified in Section \ref{sec:gripper} and contains a weak rubber band as a return spring to bring the grippers towards the surface. In our implementation the center of mass of the foot is close to the surface, so rapid deceleration of an upward swing of the front arms causes the foot to flip up, swinging the rear section of the foot between the pivot and the wall. This prevents the spines from catching onto the surface and results in failed landings during the phase 3-4 transition. Since a stronger return spring would oppose the jump motion, our solution is to use a long spar on each front foot, which contacts the wall early to limit the foot's rotation. The spar serves a second function: it reduces the required adhesion force needed at the spines, as modeled in Section \ref{sec:gripper}.

The realized robot properties and major components are listed in Table \ref{tab:table}. A final high-resolution trajectory optimization is solved using the updated link lengths, masses, and inertia from the CAD model (Stage 3 of co-design workflow in Fig. \ref{fig:codesign-workflow}) and then loaded onto the robot controller. 

\begin{figure*}[!b]
    \centering
    \includegraphics[width=\textwidth]{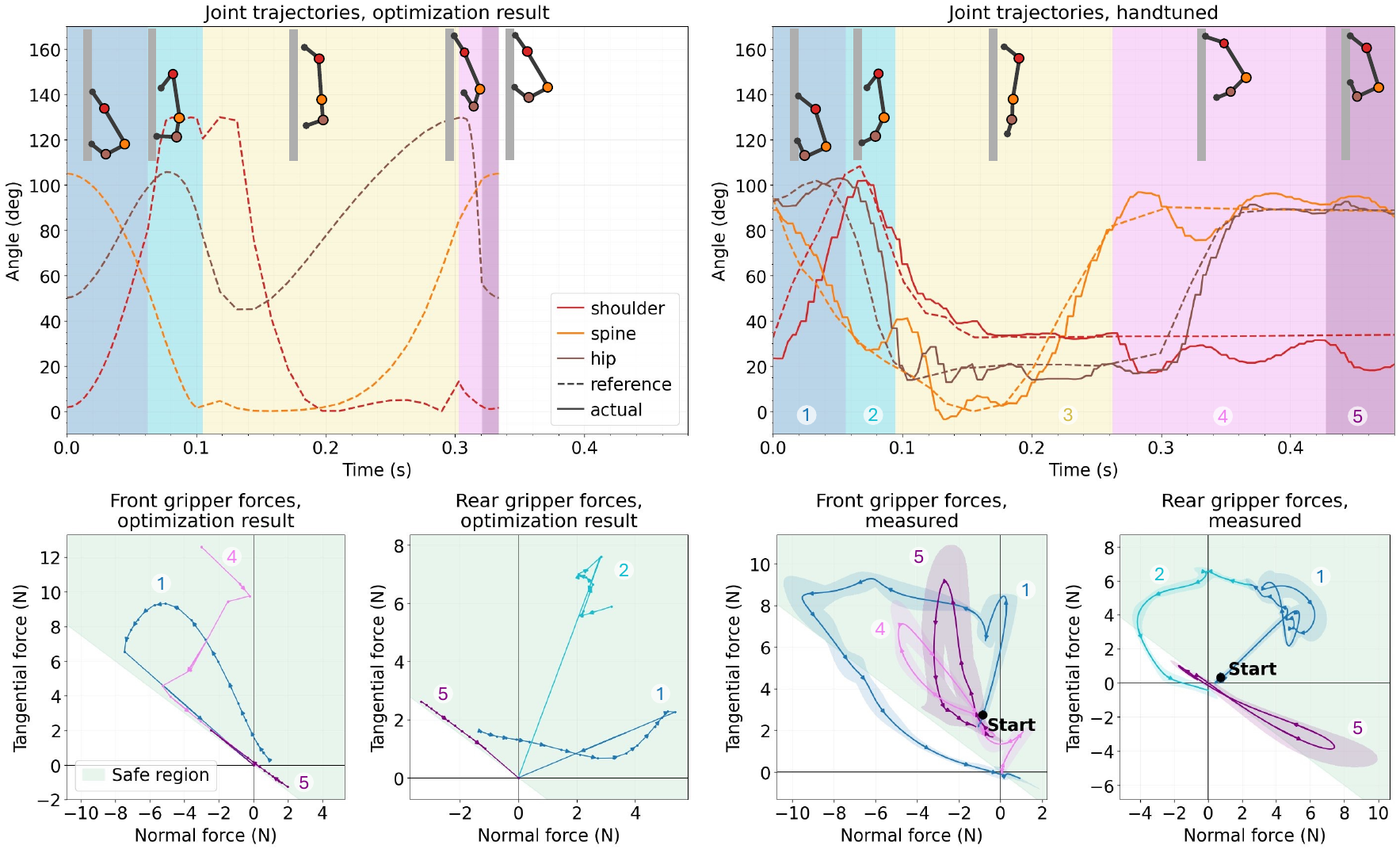}
    \caption{The optimized trajectories for the shoulder, spine, and hip joints are plotted in the top left, with phase 1-5 timings shaded. Expected gripper forces from simulation are plotted in the lower left. The manually adjusted trajectories are plotted in the upper right, along with the tracking result from a representative trial on hardware. Front and rear gripper forces were measured separately, with the average from 5 trials plotted in the lower right plots. One standard deviation is represented by the shaded region. Arrows indicate progression in time and are spaced 5 ms apart.}
    \label{fig:jumpdata}
    % \vspace{-1.25em} %if at bottom, overflows margin
\end{figure*}

\section{Experiments}

\subsection{Single jump-climb}

The joint trajectory reference from the optimization result, executed open-loop, often fails due to not reliably catching the wall with the grippers. In 30 consecutive trials up a flat surface, 12 trials caught the wall with the front grippers, but just 1 landed with both front and rear grippers such that it was ready for another jump. This may be due to imperfect tracking or errors in the reduced-order model. The joint trajectory reference was then manually tuned, keeping the fundamental design but trading off jump speed, until reliable jump-climbing is achieved. In 30 consecutive trials of the tuned trajectory, 25 trials caught the wall with both front and rear grippers. These adjustments compensated for bouncing and rapid joint oscillation after impact that the simulation did not capture. The short phase 4 of the generated trajectory was extended to allow more settling time before rear gripper contact and hip movement was simplified to reduce rapid direction changes, especially during the aerial phase that could interfere with reorientation.

We measured robot motion and contact forces for a single jump cycle on a carpeted 2D force plate (Fig. \ref{fig:region} a). Joint trajectories and 2D forces are plotted in Fig \ref{fig:jumpdata}. Joint tracking was generally underdamped but consistent across trials. Using a fitted motor model relating torque, speed, and voltage, we find that the joint torques were maximized to 0.25 Nm during pushoff (phases 1 and 2), and the shoulder, spine, and hip joints produced 0.43 J, 0.61 J, and 0.33 J of mechanical work, respectively. Gripper forces stayed within the safe region except for the front gripper at the end of phase 1 and the rear gripper at the end of phase 2, immediately before each gripper releases from the surface and force goes to zero. Brief forces outside the safe region are likely due to the dynamics of the spines unhooking from the carpet and an overly conservative safe region approximation. In both the optimization result and the experiment, the front gripper provides high adhesion and upwards force during takeoff and landing, while the rear gripper mainly provides an upwards and positive normal force during takeoff. Gripper forces during landing in the optimized result apply only adhesion forces, likely due to being timed perfectly at the jump apex, but during experiments both grippers bounce during impact and exert a brief positive normal force, more pronounced in the rear gripper. This effect is also seen as shoulder and spine joint oscillation during phases 4 and 5.

\subsection{Continuous jump-climbing and mixed locomotion}

By repeating the tuned reference trajectory, the robot is capable of continuous jump-climbing on both flat and cylindrical vertical surfaces. We measure the average speed over 4 consecutive jumps up a 12-inch diameter tube to be 0.375 m/s, or 1.97 body lengths/s.  

%(maybe plot of vertical position over time? Will show periods of very fast movement, then a lot of the time is spent stabilizing itself (phases 4 and 5))

\begin{figure}[t]
    \centering
    \includegraphics[width=\columnwidth]{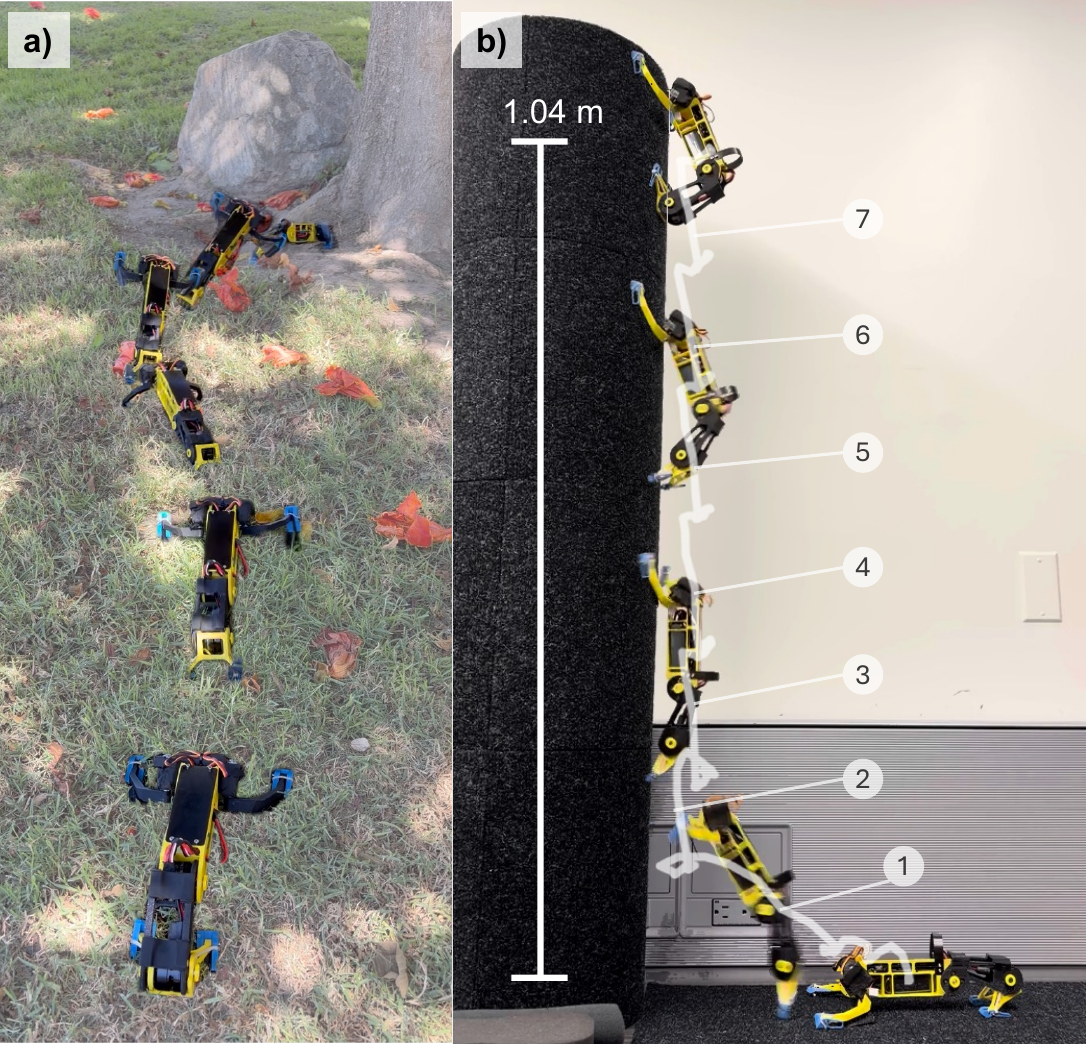}
    \caption{a) The robot traverses bumpy natural terrain. b) The robot jumps and perches to a vertical tube on jump 1, then takes six bounds upwards to raise its center of mass by 1.04 meters. A central point on the robot is tracked in white.}
    \label{fig:transition}
    \vspace{-1.25em}
\end{figure}

The same joint trajectory results in forward jumps on a variety of horizontal surfaces, including carpet, grass, and hard soil  (Fig \ref{fig:transition}a). With modifications to the reference trajectory, the robot can transition to a vertical surface in a single jump and perch passively with its spined grippers like \cite{xu2025pinto}. Repeated runs of the jump-climb trajectory then commands the robot to scale the carpeted cylinder (Fig \ref{fig:transition}b).

% \begin{figure}
%     \centering
%     \includegraphics[width=\columnwidth]{Figures/torquespeed.png}
%     \caption{}
%     \label{fig:torquespeed}
%     % \vspace{-1.25em}
% \end{figure}

% \begin{figure}
%     \centering
%     \includegraphics[width=\columnwidth]{Figures/motorwork.png}
%     \caption{}
%     \label{fig:motorwork}
%     % \vspace{-1.25em}
% \end{figure}

\section{Discussion}

We present an untethered vertical jump-climbing robot designed using a trajectory-morphology co-design workflow, as a step towards animal-like agility in vertical environments. The resulting morphology offers design insight for future serial link robot designs for climbing in the sagittal plane: a concentrated central mass, short distal links, and a relatively long reach in front of the central mass. %for reorienting the grippers to land.

\begin{figure}[t]
    \centering
    \includegraphics[height=3.5cm]{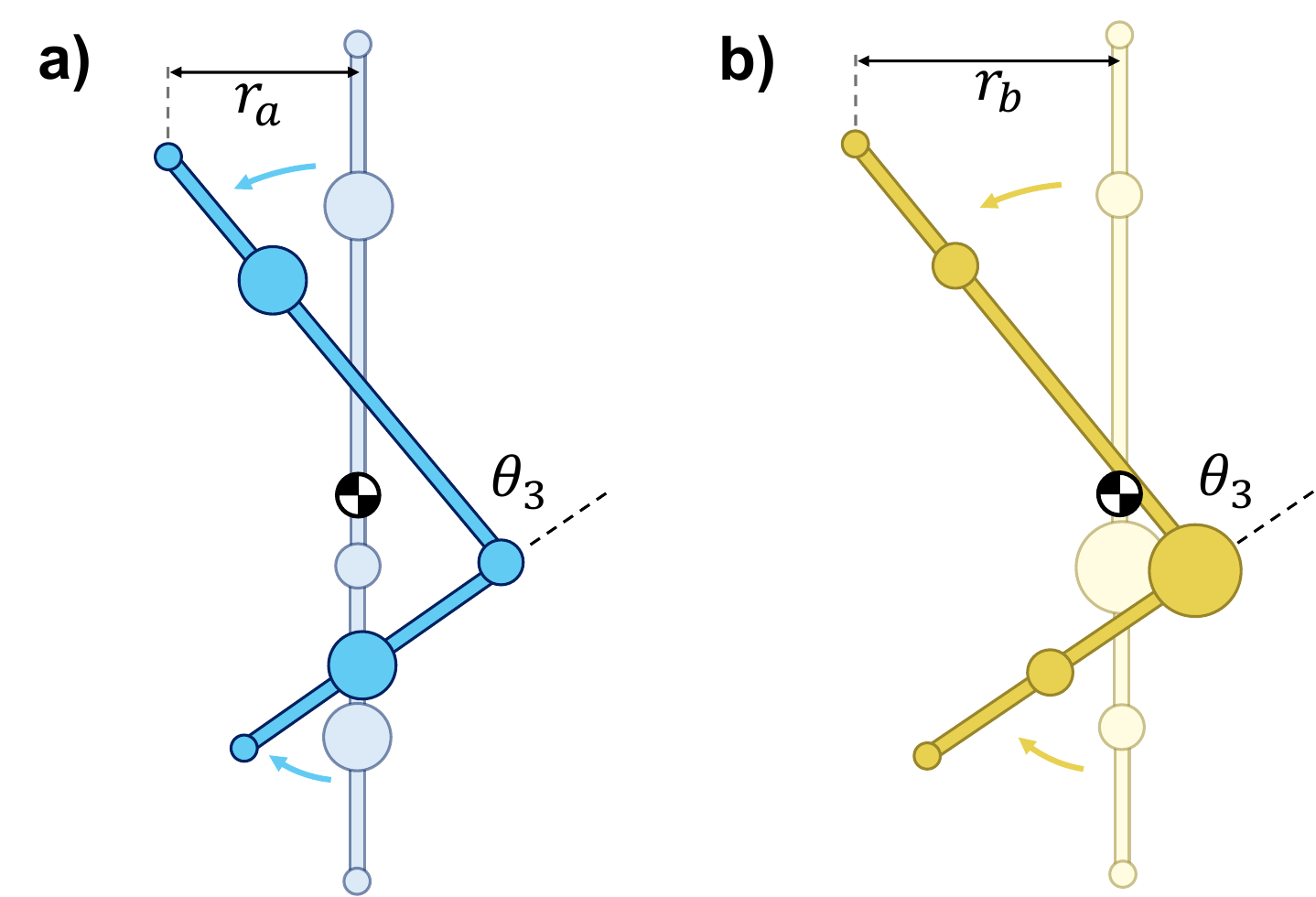}
    \vspace{-0.5 em}
    \caption{Schematic representation of the effect of mass distribution on aerial reconfiguration using the rear leg as an inertial tail. For visual simplicity, all links begin extended and only the spine joint actuates to a fixed angle $\theta_3$. Case a) has less mass at the spine and more mass at the shoulder and hip, resulting in shorter front gripper reach $r_a$ beyond the center of mass. Case b) concentrates mass at the spine and has further reach ($r_b > r_a$).}
    \label{fig:reorientation}
    \vspace{-1.25em}
\end{figure}

We suggest a few reasons that may have led to these trends. Concentrating mass at the spine and longer front links may help with better climbing performance due to being able to reach farther towards the surface with its front foot during flight (Fig. \ref{fig:reorientation}). Additionally, more mass at the spine leaves lighter limbs to accelerate during each jump. Shorter distal links may be motivated by motor torque constraints: a smaller moment arm results in higher force at the gripper for a given motor torque, and is beneficial in our case with a direct-drive motor, which was selected for its impact resistance. Finally, a shorter distance to the wall creates less pitch-back moment on the robot body and reduces the total adhesion necessary, analogous to the analysis on a single gripper in Section \ref{sec:gripper}. 

%The trajectory and morphology are optimized for climbing, and we find that the same parameters happen to generate ground locomotion. Future work could additionally consider ground locomotion performance during the optimization by constructing a weighted average of objectives for multiple independent behaviors.
% We find in separate motor characterization experiments that our motor torque is likely limited by magnetic saturation, not battery or ESC limits.

Subsequent versions can design actuators around utilization seen in our data. During a jump-climb, the spine motor produced the most work, and joint torque was saturated while speed remained under 60 rad/s, far lower than the motor speed capability at the available voltage.  For higher climbing performance, more actuator mass can be allocated to the spine joint, and speed can be traded off for higher torque using gear reductions, linkage mechanisms, or larger motors. 

To advance towards a squirrel-like robot for leaping among branches within and between trees, we should address several limitations. First, our trajectory optimization process is sensitive to errors in the dynamics model and requires manual adjustments. The adjusted open-loop trajectory spends about 45\% of a jump period settling into a state suitable for the following jump (phases 4 and 5), which reveals potential for faster climbing by reducing this settling time. Second, the spines are limited to carpet, and the trajectory optimization is dependent on force constraints characterized for this particular gripper/substrate combination. More versatile grippers should be considered for a wider range of natural surfaces \cite{spenko2023attachment}. Third, due to a lack of steering, the robot tends to veer to one side when jumping up a planar surfaces. Feedback control using an inertial measurement unit and onboard gripper force sensors would help compensate for these unmodeled effects and regulate orientation. To improve real-world transfer, the co-design workflow could evaluate control policies for each morphology in a high fidelity 3D rigid body simulation rather than using simple 2D dynamics. 

%Finally, jump-climbing is much less reliable than conventional climbing, so designers should consider the tradeoffs between agility and safety like for hopping 

% settling time between jumps, 
% spines limited to carpet, 
% open-loop control, 
% flat surfaces, 
% etc.

%The actuators can be better adapted to the application. The direct-drive brushless motors on the robot operate in a low speed and high torque regime relative to their capability.

Beyond advancing squirrel-like robots specifically, we believe this and our future work could have broader implications. Similar to how early hoppers 40 years ago helped establish design and control principles for dynamic ground locomotion with aerial phases \cite{raibert1986legged}, further exploration of maneuvers such as the jump-climb may reveal principles for dynamic vertical locomotion of more agile legged robots.
In addition, these principles could inform robophysical models of arboreal animal locomotion \cite{aydin2019robophysics}.%We can leverage co-design to explore this design space. 

%Draw parallel between dynamic climbing and dynamic hopping in Raibert hoppers. Enables further understanding of running dynamics, hopefully this work opens up opporunities to explore legged climbing. 

%One benefit of dynamic climbers is lower DOF? Cite pendular climbers and parkourbot, compare to LORIS, LEMUR, MARVEL, etc

% Perching drones extend runtime by latching onto vertical surfaces when idle, but are not intended for locomotion along the surface \cite{LussierDesbiens2011Perching}. SCAMP combines a quadrotor with a two tendon-driven arms with spined grippers, allowing it to fly, perch, and climb on rough outdoor walls \cite{pope2017scamp}. However, these systems are not intended for locomotion along the surface.

\clearpage

\section*{Acknowledgements}
We thank Alejandro Cifuentes and Xiaoyang Yu for assisting with gripper force data collection. OpenAI Codex assisted in writing firmware for robot control and experimental data collection, as well as plotting code for initial versions of figures 4-8. All generated code was tested, reviewed, and modified by the authors. 
% \raggedbottom
\bibliographystyle{IEEEtran}
\bibliography{references.bib}

\end{document}